\pdfoutput=1

\documentclass[11pt]{article}

\usepackage[]{eacl2023}

\usepackage{times}
\usepackage{latexsym}
\usepackage{enumitem}
\usepackage{multirow}
\usepackage{multicol}
\usepackage{amsmath}
\usepackage{xcolor}
\usepackage{graphicx}
\usepackage{array}
\usepackage{booktabs}
\usepackage{subcaption}
\usepackage{soul}
\usepackage{enumitem}

\usepackage[T1]{fontenc}

\usepackage[utf8]{inputenc}

\usepackage{microtype}

\title{Towards More Efficient Insertion Transformer \\with Fractional Positional Encoding}

\author{
Zhisong Zhang$^1$\thanks{~~Work done during an internship at Microsoft Research.},\quad\quad Yizhe Zhang$^2$\thanks{~~Work done at Microsoft Research.},\quad\quad Bill Dolan$^3$\\
$^1$Carnegie Mellon University, \quad\quad $^2$Apple Inc.,\quad\quad $^3$Microsoft Research\\
\normalsize{\texttt{zhisongz@cs.cmu.edu,yizhe.zhang@hotmail.com,billdol@microsoft.com}}
}

\begin{document}
\maketitle
\begin{abstract}
    Auto-regressive neural sequence models have been shown to be effective across text generation tasks. However, their left-to-right decoding order prevents generation from being parallelized. Insertion Transformer \cite{stern2019insertion} is an attractive alternative that allows outputting multiple tokens in a single generation step. Nevertheless, due to the incompatibility between absolute positional encoding and insertion-based generation schemes, it needs to refresh the encoding of every token in the generated partial hypothesis at each step, which could be costly. We design a novel reusable positional encoding scheme for Insertion Transformers called Fractional Positional Encoding (FPE), which allows reusing representations calculated in previous steps. Empirical studies on various text generation tasks demonstrate the effectiveness of FPE, which leads to floating-point operation reduction and latency improvements on batched decoding.
\end{abstract}

\section{Introduction}

Transformer-based models \citep{vaswani2017attention} have been successfully applied to various text generation tasks \citep{gong-etal-2019-enhanced,wang-etal-2019-learning-deep,ahmad-etal-2020-transformer,zhang-etal-2020-dialogpt,lewis-etal-2020-bart,brown2020language}. 
Most of these models utilize a fixed left-to-right auto-regressive generation strategy, where the strict factorization means that the model can only generate one token per step. This makes it difficult to parallelize the decoding process, while parallel generation may help to improve decoding efficiency.


Recently, insertion-based sequence-generation models \cite{stern2019insertion,gu-etal-2019-insertion,welleck2019non} have been developed as attractive alternatives to the auto-regressive ones by allowing flexible generation order. In particular, the Insertion Transformer \cite{stern2019insertion}, which combines the Transformer architecture and the insertion-based strategy, can match the performance of an auto-regressive model while requiring many fewer decoding steps with parallel generation. 

The original Insertion Transformer utilizes absolute positional encoding as in the vanilla auto-regressive transformer. In the vanilla transformer, due to its left-to-right generation scheme, tokens' absolute positions do not change; thus, previous computation can be reused.
However, this property no longer holds if insertion is allowed, and the Insertion Transformer re-encodes all previously generated tokens at each decoding step, which brings additional computational overheads.

In this work, we propose a reusable positional encoding scheme called Fractional Positional Encoding (FPE) to accelerate the Insertion Transformer. This scheme dynamically calculates each token's positional representations according to its left and right neighbors at insertion time. In this way, each token's positional representations will not change during the decoding process so that the computation can be reusable in the same way as the vanilla transformer, leading to a reduction of computation for the Insertion Transformer.

\begin{figure*}[t]
	\small
	\centering
	\includegraphics[width=0.975\textwidth]{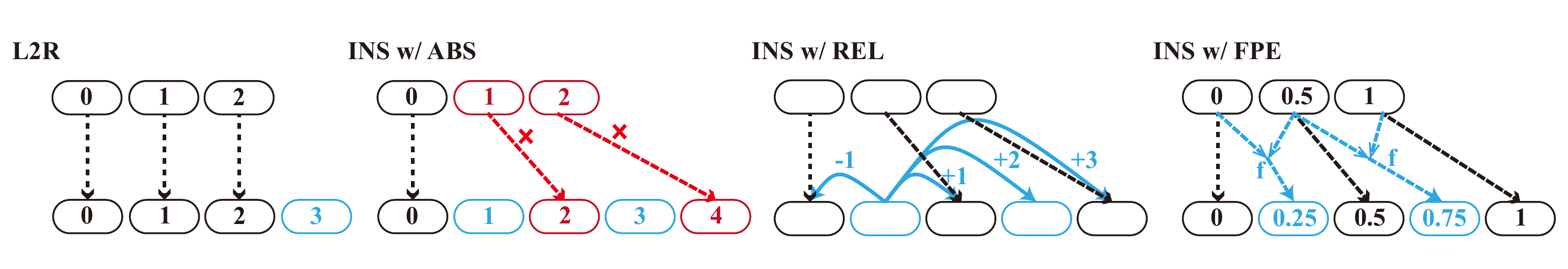}
	\caption{Illustrations of different positional encoding schemes. Black (or red) nodes denote the tokens that already exist in the previous step, while the newly generated ones are in blue. In ABS, the positional embeddings in previous steps may be non-reusable since absolute positions may change (marked as red). Representations can be made reusable by using REL or our proposed FPE. (Note that the fractional numbers in the FPE figure are only for better illustration, in practice, we adopt embeddings calculated by a learnable linear function.)}
	\label{fig:posi}
\end{figure*}

We evaluate FPE with a range of text generation tasks, including machine translation, word reordering, summarization as well as an open-ended text completion task. We show that the proposed scheme can reduce floating point operations of the insertion-based model while maintaining comparable performance to the vanilla transformer.

Our implementation is available at \url{https://github.com/zzsfornlp/zgen1/}.


\section{Insertion Transformer}


Insertion Transformer \cite{stern2019insertion} generates the target sequences via a series of insertion operations. This provides a flexible scheme that can enable different generation orders as well as parallel generation. We focus on the parallel-generation variant that inserts multiple tokens at each step. While the left-to-right scheme can only append one token at each step, the insertion-based scheme can add multiple tokens at different slots, thus potentially enabling more efficient generation.

\section{Positional Encoding}

Figure~\ref{fig:posi} provides an overview of different positional encoding schemes that we explore.
The vanilla left-to-right (\textbf{L2R}) Transformer model \cite{vaswani2017attention} adopts a simple absolute positional encoding scheme by assigning left-to-right increasing indexes to each token. 
This naturally fits the left-to-right generation procedure and allows the previously calculated hidden representations to be reusable.
However, in the insertion-based generation, since tokens can be inserted before previously generated tokens, the absolute position of a token may change. Therefore, if still using the absolute positional encoding (\textbf{ABS}), the previously calculated hidden layers cannot be reused and the Insertion Transformer needs to re-encode all the existing tokens at each step. This yields computation overhead, which may offset the computation gain from parallelization.

To solve this problem, alternative positional encoding schemes are required. Relative positional encoding \citep[\textbf{REL};][]{shaw-etal-2018-self} is an example, which has been adopted for insertion-based models \citep{lu2021efficient}. Here, each token records its relative positional information at its insertion time.
Though this scheme alleviates encoding absolute positions and allows reusing, it requires complex modifications in the attention calculations.

In this work, we design fractional positional encoding (\textbf{FPE}), which is a simpler alternative scheme that only modifies the input embeddings.
We still give each token a positional embedding $\mathbf{p}$, which is dynamically calculated along the generation process.
Whenever a new token $w_{\text{new}}$ is inserted between two existing tokens $w_{\text{left}}$ and $w_{\text{right}}$, its positional representations will be calculated with a function $f$ applying to its current left and right neighbors: $\mathbf{p}_{\text{new}} = f(\mathbf{p}_{\text{left}}, \mathbf{p}_{\text{right}})$.
In this way, we will have the ``fractional''-styled positions. The positional representations of all the tokens will not change throughout the decoding process, and re-encoding is no longer needed.

\begin{table*}[t]
	\centering
	\small
	\begin{tabular}{l | c | c | c | c | c }
		\toprule
		Task & Model & Evaluation~$\uparrow$ & \#Step & \#Len & Latency~$\downarrow$  \\
		\midrule
		\multirow{4}{*}{\shortstack[l]{Translation \\ (WMT14 EN-DE)}} & L2R & \textbf{27.72} & 28.4 & 22.1 & 230.1 \\
		& ABS & 27.45 & 5.7 & 21.5 & 100.3 \\
		& REL & 27.40 & 5.5 & 21.5 & 105.0 \\
		& FPE & 27.47 & 5.6 & 21.4 & \textbf{97.2} \\
		\midrule
		\multirow{4}{*}{\shortstack[l]{Text Reordering \\ (Wiki-103)}} & L2R & \textbf{52.82} & 27.9 & 24.8 & 224.7 \\
		& ABS & 50.69 & 6.8 & 24.2 & 113.9 \\
		& REL & 52.63 & 6.0 & 24.6 & 120.7 \\
		& FPE & 52.52 & 6.0 & 24.8 & \textbf{105.9} \\
		\midrule
		\multirow{4}{*}{\shortstack[l]{Summarization \\ (XSUM)}} & L2R & 31.33/11.65/25.32 & 21.3 & 19.7 & 206.9 \\
		& ABS & \textbf{32.09}/11.39/\textbf{25.68} & 6.7 & 24.3 & 114.9 \\
		& REL & 31.90/\textbf{11.66}/25.80 & 6.2 & 22.9 & 125.2 \\
		& FPE & 31.78/11.57/25.67 & 6.2 & 22.7 & \textbf{114.4} \\
		\midrule
		\multirow{4}{*}{\shortstack[l]{Text Completion \\ (Wiki-103)}} & L2R & \textbf{3.87}/\textbf{8.48}/\textbf{14.54} & 55.6 & 48.9 & 468.0 \\
		& ABS & 1.19/7.66/12.90 & 9.5 & 49.4 & 141.9 \\
		& REL & 1.69/8.41/13.23 & 8.1 & 54.7 & 161.0 \\
		& FPE & 1.61/8.26/13.47 & 8.0 & 52.3 & \textbf{129.9} \\
		\bottomrule
	\end{tabular}
	\caption{Main results of comparing an auto-regressive left-to-right (L2R) model and three insertion models with absolute (ABS), relative (REL), and fractional (FPE) positional encoding. ``Evaluation'' denotes automatic evaluation metrics: BLEU for MT and reordering, R-1/R-2/R-L for summarization, and BLEU/METEOR/R-L for completion. ``\#Step'' and ``\#Len'' indicate the average decoding steps and output lengths, respectively. ``Latency'' denotes the actual average decoding time (ms) per instance with single-instance decoding.}
	\label{tab:mainc}
\end{table*}

We specify the FPE representations $\mathbf{p}$ to have the same dimension as the model size and add them to the input embeddings as in the vanilla transformer. We further specify two randomly-initialized embeddings $\mathbf{p}_B$ and $\mathbf{p}_E$ for the beginning- and ending- of sequence tokens, respectively. The function $f$ is modeled by a linear layer\footnote{We start with the simple linear layer and find it works reasonably well. We also tried some other methods such as adding non-linearity activation but did not find obvious benefits. Therefore, we adopt this simple method.} which takes the concatenation of the two neighbors' positional embeddings and outputs a new vector of the model size. At training time, all these FPE-related parameters are tuned along with other parameters in the model. The linear layer is lightweight compared to the transformer layers, thus introducing negligible cost.

\section{Experiments}

\subsection{Settings}

We explore a variety of generation tasks, including machine translation, word reordering, summarization, as well as an open-ended text completion task. We use WMT14 En-De \cite{bojar-etal-2014-findings} for machine translation, sentences in WikiText-103 \citep[Wiki103;][]{merity2016pointer} for word reordering, XSUM \cite{narayan-etal-2018-dont} for summarization and paragraphs in Wiki103 for completion. 
In the text completion task, the model is required to complete each paragraph according to the existing context. 
All the decoding experiments are performed with one V100 GPU.
Please refer to Appendix \ref{app:data} and \ref{app:exp} for more dataset and experimental details.

\subsection{Results}

\begin{figure}[t]
	\small
	\centering
	\includegraphics[width=0.49\textwidth]{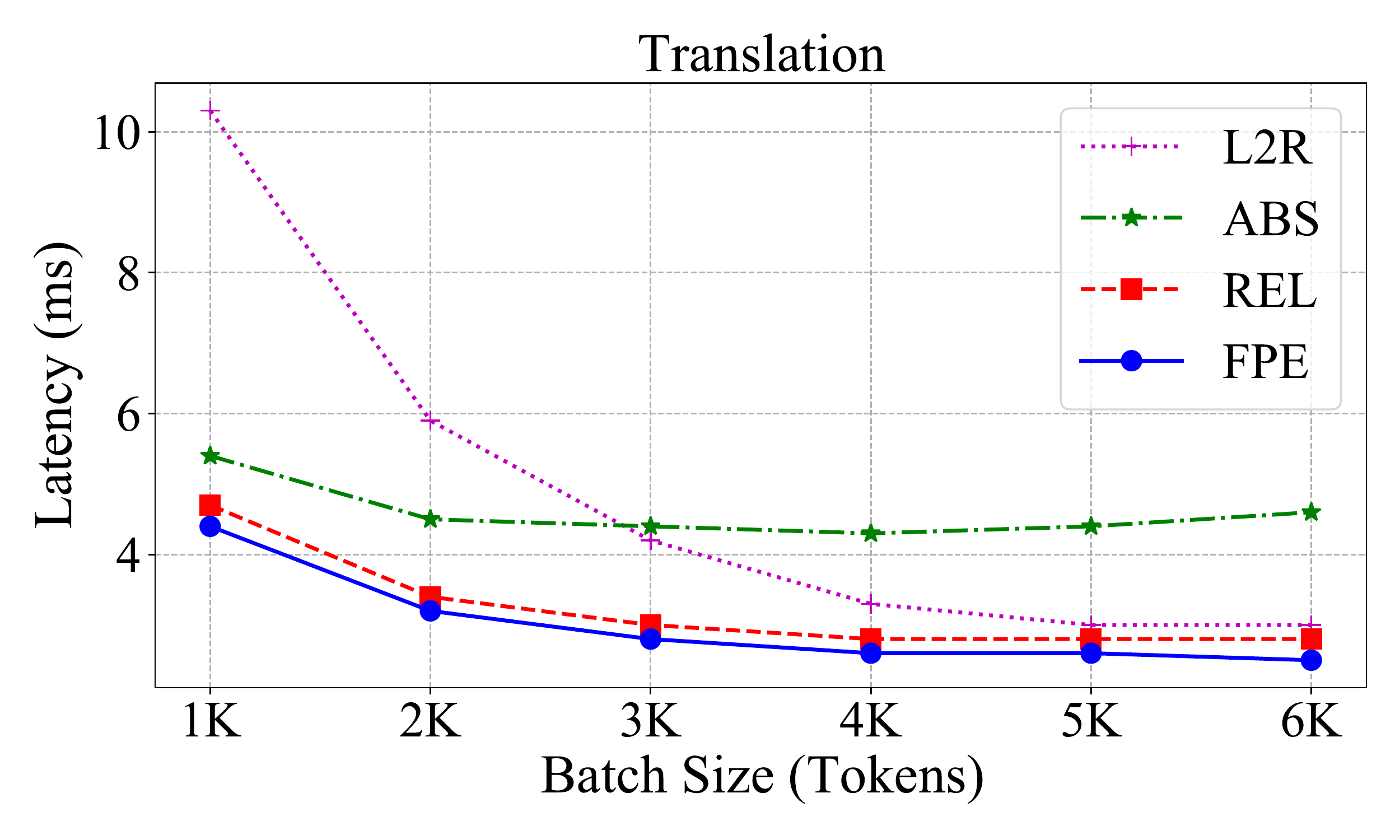}
	\caption{Latency of MT models with different decoding batch sizes (source tokens). Results with single-instance decoding are not shown here since since its latency is much higher.}
	\label{fig:batch1}
\end{figure}

\begin{table}[t]
	\centering
	\small
	\begin{tabular}{c|c|c|c|c|c|c}
		\toprule
		Batch-size & 1K & 2K & 3K & 4K & 5K & 6K \\
        \midrule
L2R & 10.3 & 5.9 & 4.2 & 3.3 & 3.0 & 3.0 \\
ABS & 5.4 & 4.5 & 4.4 & 4.3 & 4.4 & 4.6 \\
REL & 4.7 & 3.4 & 3.0 & 2.8 & 2.8 & 2.8 \\
FPE & 4.4 & 3.2 & 2.8 & 2.6 & 2.6 & 2.5 \\
		\bottomrule
	\end{tabular}
	\caption{Latency of MT models (milliseconds per instance) with different decoding batch sizes (source tokens). This table shows the detailed numbers corresponding to those in Figure~\ref{fig:batch1}.}
	\label{tab:batch1}
\end{table}

We compare our method (FPE) with the vanilla transformer (L2R), and two other insertion-based models with absolute (ABS) and relative (REL) positional encoding.
The main results are shown in Table~\ref{tab:mainc}.
For automatic performance evaluations, the three insertion-based transformer models (ABS, REL, and FPE) achieve similar results. Compared with L2R, the insertion models' performance is competitive on MT, reordering, and summarization tasks, while being behind on the open-ended text completion task. This is presumably due to the conditional independence assumption in the parallel generation steps. 
This issue is beyond the scope of this paper, so we leave it to future work. 

For efficiency, insertion-based models can generate target sequences with much fewer decoding steps, leading to latency reduction where the insertion models can achieve around 2x speedups compared to L2R in single-instance mode.

\subsection{Batched Decoding}

We further explore batched decoding, which is usually adopted to speed up the computation via data parallelism. The latency of MT models against different batch sizes can be found in Figure~\ref{fig:batch1} and Table~\ref{tab:batch1}, from which we observe that:
\begin{itemize}[noitemsep,leftmargin=*]
\item ABS becomes less efficient when decoding in batches, probably due to the extra computations\footnote{We further measure the floating point operations (FLOPs) for decoding an instance following the method of \citet{Elbayad2020Depth-Adaptive} and find that the ABS-based model requires much larger FLOPs than other models. For example, on the MT models, FLOPs per instance is 8.69B for ABS, while FPE only requires 4.65B (REL needs 4.68B).} brought by re-encoding.
Though this does not affect its efficiency in the single-instance mode where GPU's computational capacity may not be fulfilled yet, in batched decoding the extra re-encoding computations greatly dampen its efficiency improvements.
\item FPE and REL are faster than L2R for relatively small batch sizes. While the batch size becomes larger, the efficiency gain becomes less obvious. Presumably, more complex indexing operations in the insertion-based schemes do not utilize GPUs as fully as L2R. We leave this optimization to future work. 
\item REL behaves similarly to FPE, but is consistently around 10\% slower, probably due to the extra relative positional computations in attentions.
\end{itemize}
The patterns in other tasks are similar to MT and are shown in Appendix \ref{app:res}.


Note that many previous works consider only single-instance or batched decoding mode when measuring efficiency, while we examine both to include a spectrum of real scenarios covering various device memory capacities and querying patterns. 
While the L2R model and the original Insertion Transformer (w/ ABS) only excel at one end, FPE could help to make the model efficient for both scenarios, potentially benefiting more use cases.

\begin{figure}[t]
	\small
	\centering
	\includegraphics[width=0.49\textwidth]{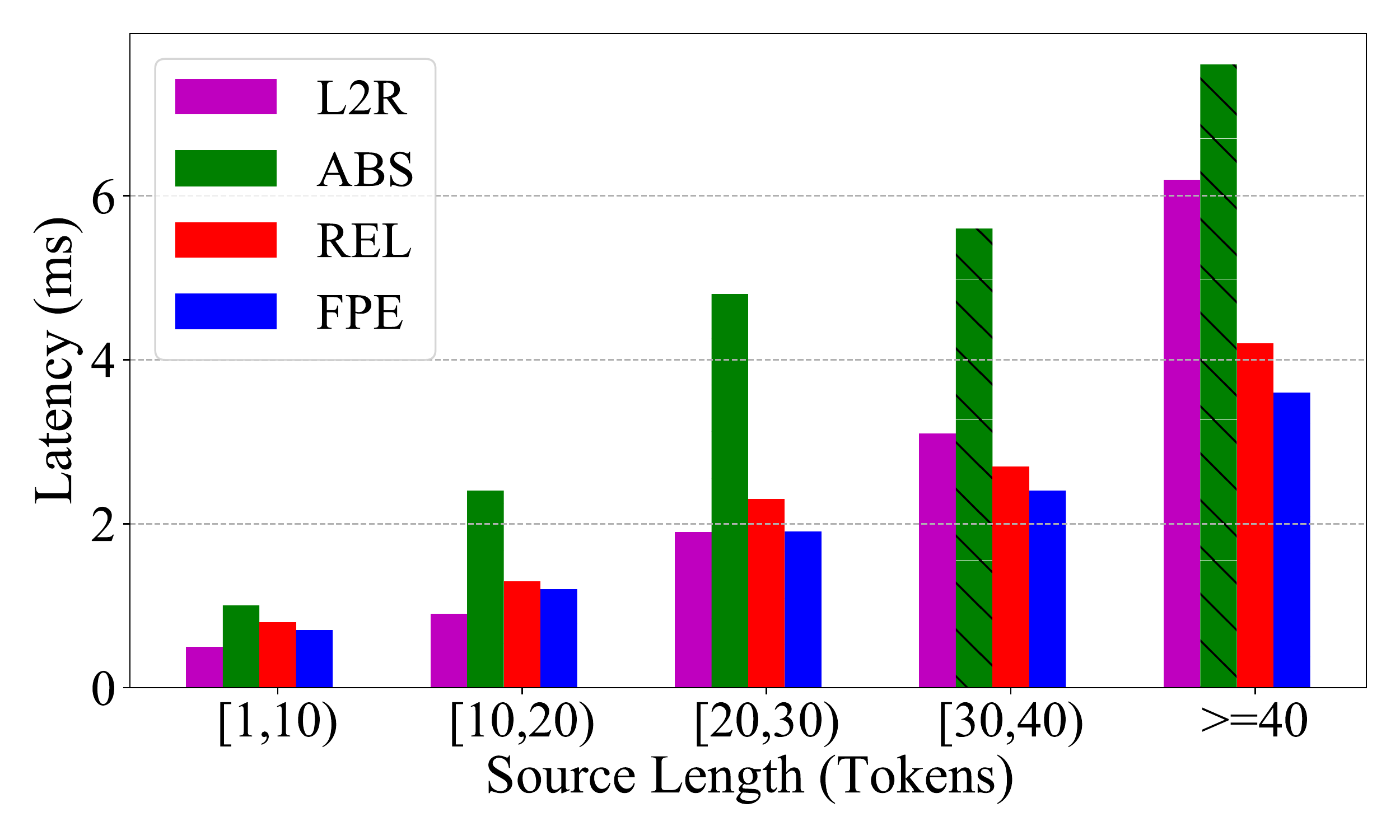}
	\caption{Latency breakdowns on source sequence lengths for the MT task with batched decoding (6K).}
	\label{fig:bins}
\end{figure}

\subsection{Length Breakdown} 
We further perform an ablated speed analysis by breaking down the instances by sequence length. Here, we investigate the task of machine translation and split the instances into different bins according to source lengths. The breakdown results with a batch size of 6K are shown in Figure~\ref{fig:bins}. 
The trends are generally similar to the overall results. ABS is not quite efficient for batched decoding and can be even slower than L2R. Though the utilization of FPE does not provide obvious efficiency improvements on short sentences, it brings benefits for generating longer sequences. It can achieve a 1.7x speedup over L2R for sentences that are longer than 40 tokens, even with a large batch size. Similar to the overall trend, FPE is consistently more efficient than REL, yet being simpler.

\section{Related work}

\paragraph{Generation Order.}
Previous works have been exploring relaxing the output dependencies and allowing parallel generation. The Non-Autoregressive Transformer (NAT) \cite{gu2018non} enables the decoder to generate target sequences in one or several decoding steps \cite{gu2018non,gu-kong-2021-fully,lee-etal-2018-deterministic,NEURIPS2019_675f9820,ghazvininejad-etal-2019-mask}. Most of these models require predicting target length and generating multiple consecutive tokens at once. The generation is sometimes not fluent, as multiple tokens may compete for the same meaning. 
The insertion-based methods \cite{stern2019insertion,gu-etal-2019-insertion,welleck2019non,chan-etal-2020-empirical,zhang-etal-2020-pointer} also change the standard left-to-right generation by allowing dynamically inserting tokens for the generation process. This provides a good balance between generation fluency and efficiency, and does not require predicting target lengths first.
In this work, we follow this insertion-based generation scheme and further improve its efficiency. 
In addition to efficiency, allowing flexible generation order is another motivation to study non-L2R generation schemes. A good generation order may bring performance benefits \citep{ford-etal-2018-importance,jiang-bansal-2021-learning-analyzing}.


\paragraph{Reusable Positional Encoding.}
In insertion-based models, the absolute positional encoding can be non-reusable since an inserted token will change the absolute positions of its following tokens. Alternative schemes are required to enable reusable encoding.
Relative positional encoding \cite{shaw-etal-2018-self} is utilized for the insertion-based generation in some recent work \cite{gu-etal-2019-insertion,lu2021efficient} to avoid the re-encoding of the previously generated tokens. However, it requires modifications to the inner attention mechanism of the Transformer, while our FPE scheme is a simpler alternative that only modifies the input. 
In a similar spirit to our scheme, \citet{shiv2019novel} explore a tree-based positional encoding strategy. Our scheme is different in that in the insertion-based generation, each node has two parent nodes, yielding a graph structure rather than a tree.

\section{Conclusion}

In this work, we investigate the re-encoding issue that sometimes hinders the Insertion Transformer from receiving its computation gain and propose a Fractional Positional Encoding scheme that is naturally compatible with the insertion-based generation scheme to solve this issue. With experiments on various tasks, we show that this simple scheme eliminates the need of re-encoding the previously generated tokens and obtains a promising balance between efficiency and performance.

\section*{Limitations}

This work has several limitations. First, we mainly rely upon the architecture and decoding strategy of the Insertion Transformer, which only allows the generation of one token between two neighboring tokens at one step. It would be interesting to consider more flexible generation schemes. It would also be interesting to compare our models with other (semi) non-autoregressive models, which we leave to future work. Moreover, we follow the best-performing binary tree training objective of the Insertion Transformer, which in some way sacrifices the flexibility of the generation order. It would be interesting to explore the application of the proposed positional encoding scheme with more flexible generation orders. It would also be interesting to explore the impacts of using larger pre-trained models and investigate how it interacts with the insertion-based scheme. Finally, on the open-ended generation task, the insertion-based method still performs worse than the left-to-right one, which requires further investigation since the output dependencies would need more careful modeling in the open-ended scenarios. 

\section*{Broader Impact}
\label{app:bi}
This work focuses on improving for the natural language processing (NLP) and general artificial intelligence (AI) research community. Our work can be leveraged to improve natural language generation (NLG) models, including but not limited to text editing, conversational agents, and question answering systems.
The \textbf{broader impact} and the \textbf{risks} of this work are summarized as following:
\begin{itemize}[wide=0\parindent,noitemsep,topsep=0em]
    \item 
    This work can facilitate research in the NLG tasks in a generic manner, to potentially accelerate generations in applications like machine translation, text summarization, and virtual assistants. 
    \item This work is a fundamental research work that focuses on technical improvement, thus we have NOT imposed additional aggressive filtering techniques to the text data we used, beyond what has been performed to the original dataset from their sources. The text data we used may have offensiveness/toxicity/fairness/bias issues that we haven't been able to identify, as those are not the main focus of this work. 
    \item Given the above potential risk, due to the nature of natural language generative models, we note that the generations or outputs of this work, though not likely, may reflect gender and other historical biases implicit in the data. Under rare circumstances, the generations may exhibit a mild extent of unethical, biased, or offensive attitudes. These are known issues with current state-of-the-art text generation models. We would hope that a faster generation system as what we present can enable more iterations of further mitigation strategies to inappropriate and hallucinated generations.
    \item This work aims to advance AI technology in an environmental-friendly manner. Our proposed method can potentially reduce carbon footprints produced by AI models.
\end{itemize}

\bibliography{ins_gen}
\bibliographystyle{acl_natbib}

\newpage
\appendix

\section{Dataset details}
\label{app:data}

We provide more details of the datasets utilized in this work:

\noindent$\bullet~$\textbf{WMT14 (En-De).} For machine translation, we utilize the widely used WMT 2014 English-German translation dataset \cite{bojar-etal-2014-findings}, with newstest2013 as the development and newstest2014 as the test set. Following previous work \cite{stern2019insertion,chan-etal-2020-empirical}, we apply sequence-level knowledge distillation \cite{hinton2015distilling,kim-rush-2016-sequence} from a left-to-right autoregressive model, which has been found helpful to reduce data complexity and improve the performance of NAT models \cite{Zhou2020Understanding}.

\noindent$\bullet~$\textbf{Wiki103(S).} For word reordering, we take text sequences from the WikiText-103 dataset \cite{merity2016pointer}. Here, we focus on the task at the \textbf{S}entence level and thus perform sentence-splitting and treat each sentence as an individual sequence.

\noindent$\bullet~$\textbf{XSUM.} For summarization, we utilize the XSUM dataset \cite{narayan-etal-2018-dont}, where the targets are short, one-sentence news summaries for news articles. This task does not favor the extractive strategies and provides a good test bed for abstractive generation-based models. Following previous work \cite{liu-lapata-2019-text}, We truncate the input documents to 512 tokens.

\noindent$\bullet~$\textbf{Wiki103(P).} For paragraph completion, we again utilize the WikiText-103 dataset, but at the \textbf{P}aragraph level this time. We take paragraphs that contain four to seven sentences. For each paragraph, we take the last two sentences as the target and the previous ones are used as the source inputs.


Table~\ref{tab:data} summarizes the statistics of the datasets.

\section{Experimental settings}
\label{app:exp}

We mainly follow the settings of the original Insertion Transformer \cite{stern2019insertion}. To further encourage generations in fewer steps, we adopt the Binary Tree training loss. For the REL-based insertion model, we include the relative positional encoding by modifying attentions following \citep{shaw-etal-2018-self}. For other hyper-parameter settings, we mainly follow the common practice. We adopt slightly different settings for constrained and open-ended tasks, which are described in the following.

For the source-constrained tasks (MT, reordering, and summarization), we take the Transformer-base architecture \cite{vaswani2017attention} (6 layers, 8 heads per layer, 512 model dimensions) and the full model contains around 66M parameters. 
The models are trained by the Adam optimizer \cite{kingma2014adam}, with the same learning rate scheduling scheme of \citep{vaswani2017attention}.
We train the models for a maximum of 300K steps for machine translation and 100K steps for reordering and summarization. 
The models are validated on the development set every 1K steps and we average the five checkpoints that obtain the best results to obtain the final model. 
We take standard evaluation metrics for the corresponding tasks: BLEU\footnote{\url{https://github.com/moses-smt/mosesdecoder/blob/master/scripts/generic/multi-bleu.perl}} \cite{papineni-etal-2002-bleu} for machine translation and word reordering, ROUGE\footnote{\url{https://github.com/google-research/google-research/tree/master/rouge}} \cite{lin-2004-rouge} for summarization. Unless otherwise specified, we utilize a beam size of 4 in decoding. Following \citet{stern2019insertion} and \citet{chan-etal-2020-empirical}, we select an EOS penalty $\in$ \{0, 0.5, 1, ..., 5\} according to the results on the development set.

\begin{table}[t]
	\centering
	\small
	\begin{tabular}{c | c | c | c}
		\toprule
		Datasets & \#Seq.(train/dev/test) & Src-Len & Trg-Len \\
		\midrule
		WMT14 & 4.0M/3.0K/3.0K & 26.1 & 24.8 \\
		Wiki103(S) & 1.8M/3.8K/4.1K & 25.7 & 25.7 \\
		XSUM & 204K/11.3K/11.3K & 328.5 & 23.3 \\
		Wiki103(P) & 349K/0.8K/0.8K & 79.9 & 50.6 \\
		\midrule
		\bottomrule
	\end{tabular}
	\caption{Statistics of the datasets utilized in this work. Here, ``\#Seq.'' denotes the number of instances (sequences) in each split, ``Src-Len'' indicates the average number of words in the source, and ``Trg-Len'' shows the average number of words in the target.}
	\label{tab:data}
\end{table}

For the open-ended paragraph completion task, we adopt similar schemes, but with a difference of employing pre-trained models, which we find helpful in preliminary experiments. Due to limitation of computational resources, we adopt a relatively small model: the distilled version\footnote{\url{https://huggingface.co/distilroberta-base}} \cite{sanh2019distilbert} of RoBERTa \cite{liu2019roberta} (6 layers, 12 heads per layer, 768 model dimensions). The full model contains around 140M parameters. We adopt similar training schemes to the constrained cases and the models are trained for 300K steps. Since there are no reliable automatic evaluation metrics for this task, we simply average the final five checkpoints as the final model. For the open-ended task, we find that using greedy or beam search sometimes leads to outputs with severe repetition problems, we thus apply sampling, specifically, nucleus sampling with $p$=0.95 \cite{Holtzman2020The} in decoding.

All the models are trained with four V100 GPUs and tested with one V100 GPU. The training takes one to three days depending on the tasks. 

\begin{figure}[t]
	\small
	\centering
	\begin{subfigure}[b]{0.48\textwidth}
		\includegraphics[width=0.975\textwidth]{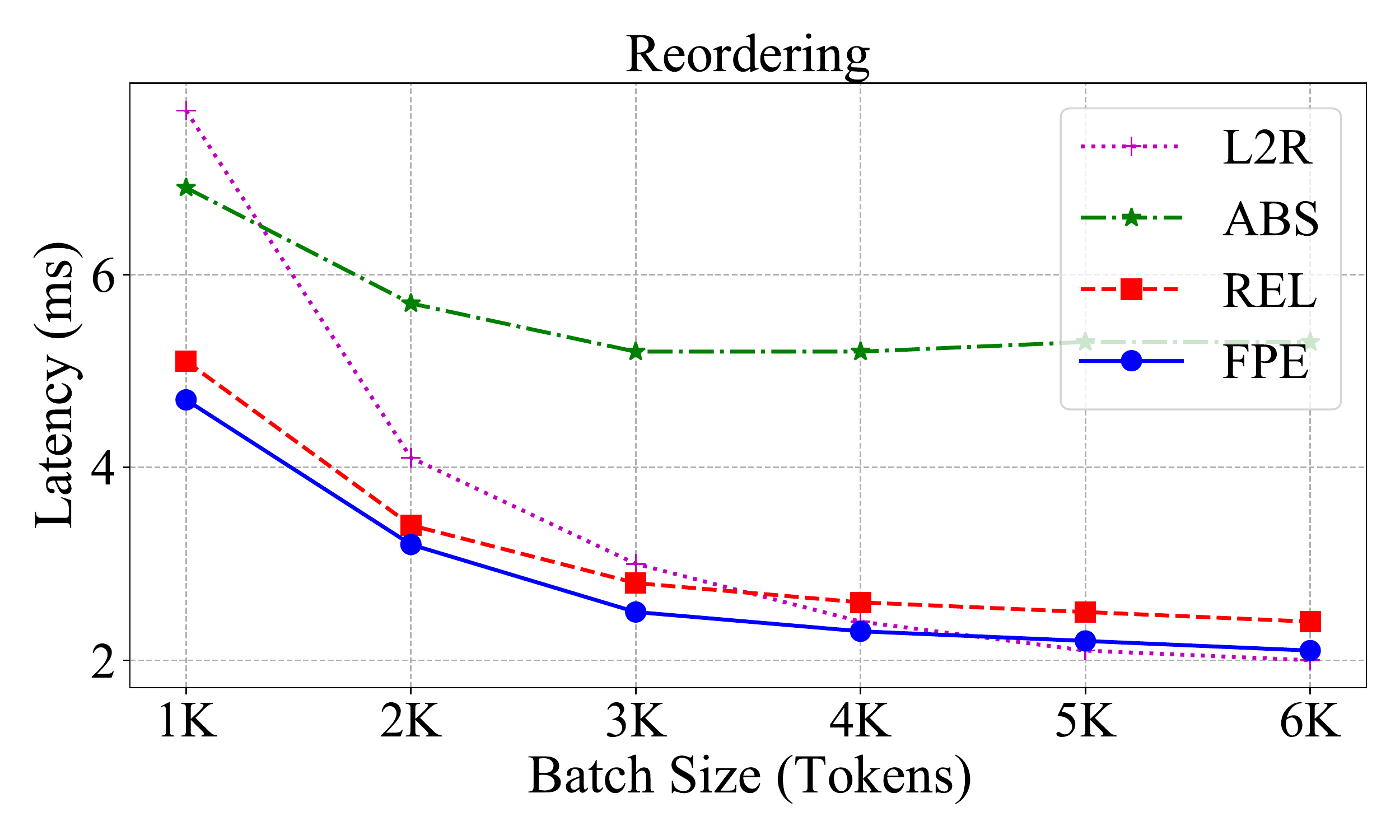}
	\end{subfigure}
	\begin{subfigure}[b]{0.48\textwidth}
		\includegraphics[width=0.975\textwidth]{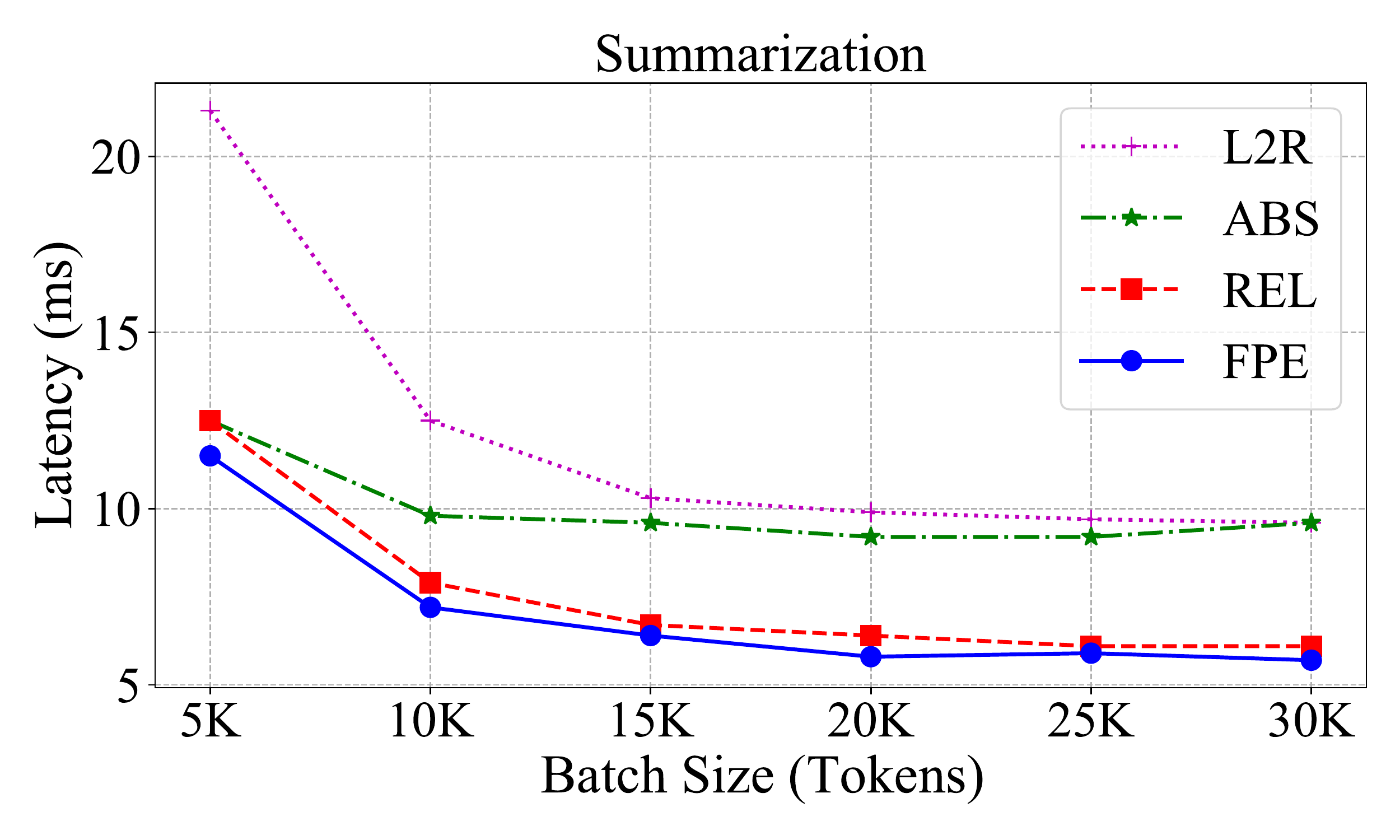}
	\end{subfigure}
	\begin{subfigure}[b]{0.48\textwidth}
		\includegraphics[width=0.975\textwidth]{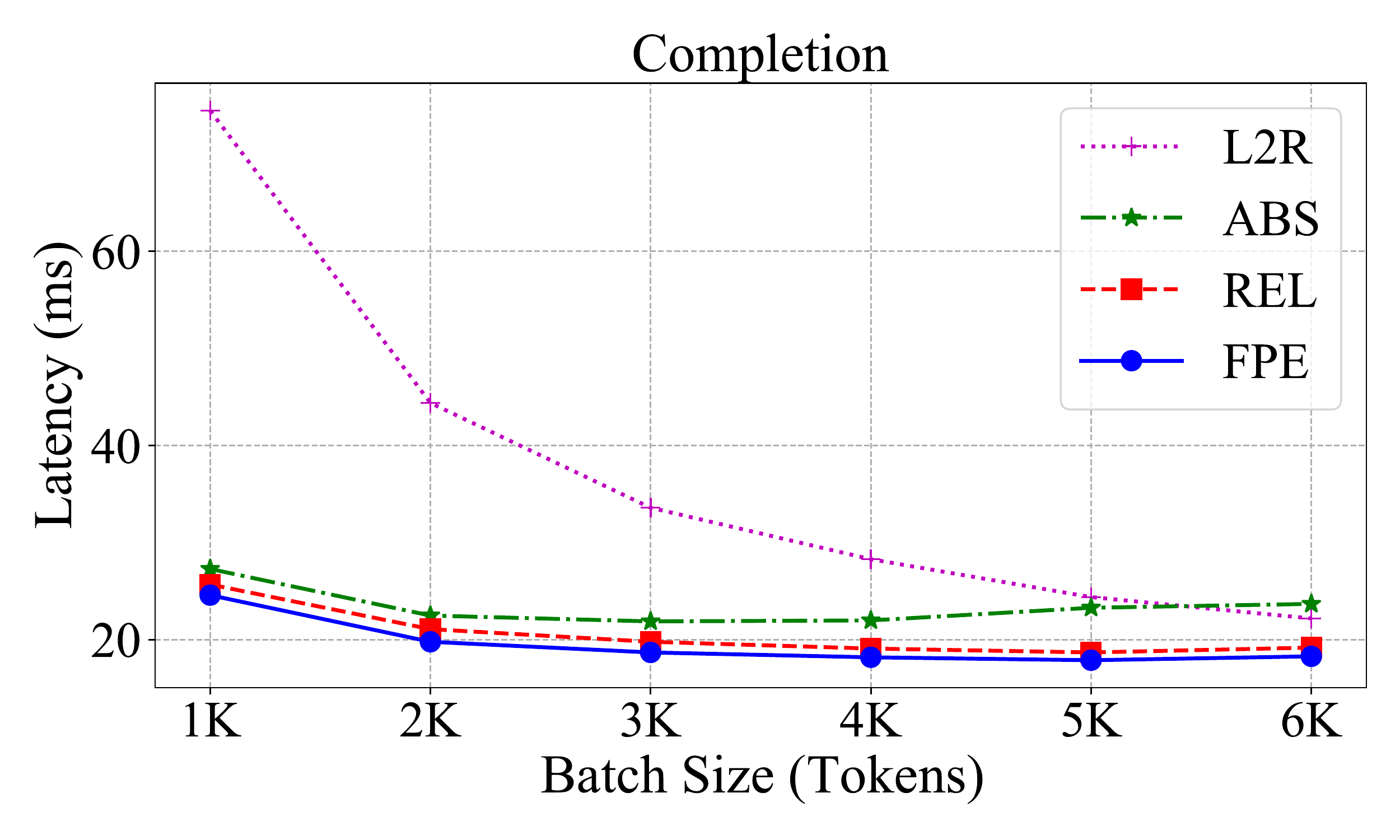}
	\end{subfigure}
	\caption{Latency of the models with different decoding batch sizes (source tokens) for more tasks.}
	\label{fig:batch2}
\end{figure}

\section{Additional Results}
\label{app:res}

In Figure~\ref{fig:batch2}, we further show the batched-decoding latency of different models on more tasks. The patterns are generally similar to those in Figure~\ref{fig:batch1} of the MT task.

\end{document}